# Deeply Explainable Artificial Neural Network

David Zucker[1]

## Abstract

While Deep learning models have demonstrated remarkable success in numerous domains, their black-box nature is still a significant limitation, especially in critical fields such as analysis and inference from medical images. Existing explainability methods—such as SHAP, LIME, and Grad-CAM—are typically applied *post hoc*, adding computational overhead and sometimes producing inconsistent or ambiguous results. In this paper, we present Deeply Explainable Artificial Neural Network (DxANN), a novel deep learning architecture that embeds explainability *ante hoc*, in the training process. Unlike conventional models that require external methods for interpretation, DxANN is designed to produce per-sample, per-feature explanations as part of the forward pass. Built on a flow-based framework, it enables both accurate predictions and transparent decision-making, and is particularly well suited for image-based tasks. While our focus in this paper is on medical images, the DxANN architecture can be readily adapted to other data modalities, including tabular and sequential. DxANN marks a step forward to intrinsically interpretable deep learning, offering a practical solution for major applications where trust and accountability are essential.

## 1. Introduction

Deep learning has revolutionized numerous fields, enabling breakthroughs in image recognition, natural language processing and biomedical analysis. Despite these successes, the opacity of neural networks has raised significant concerns and impeded adoption in critical domains where interpretability and explainability are essential. The inability to understand how a model arrives at its decisions limits trust, hinders regulatory approvals, and complicates error analysis. As a result, the field of Explainable Artificial Intelligence (XAI) has emerged.

Traditional XAI techniques are predominantly *post hoc,* which means that they rely on external methods to extract explanations from inferences and/or parameters of already trained models. Methods such as SHAP and LIME approximate feature importance, while visualization techniques such as Grad-CAM highlight influential input image regions in classification tasks. Unfortunately, the *post hoc* methods have intrinsic limitations: they depend on approximations that may not accurately reflect the true inference process, often provide inconsistent explanations for different runs or datasets, and introduce additional implementation complexity.

To address these challenges, we propose DxANN, a deep learning framework designed to provide explainability *ante hoc*, as a built-in property. The key innovation of DxANN lies in its architecture, which internally generates per-feature, per-sample explanations at inference time.

This paper describes the theoretical foundations, architectural design, and empirical performance of DxANN with two medical image datasets. We explore the ability of DxANN to maintain high classification accuracy while simultaneously producing meaningful, consistent explanations. The adaptability of DxANN to various data types, including tabular data and time-series, is also discussed.

---

[1] Arteevo Technologies Ltd.

The remainder of this paper is structured as follows: Section 2 outlines related work in explainable AI and interpretable deep learning architectures. Section 3 provides detailed mathematical formulation and design principles of DxANN. Section 4 describes our experimental setup and evaluation metric, followed by a discussion in Section 5. Conclusions, insights into the broader implications of DxANN in the field of interpretable machine learning and potential future work are discussed in Section 6.

## 2. Related work

Explainability in deep learning has been explored extensively, especially for image processing models. Gradient-based *post hoc* methods, such as Grad-CAM[2] and integrated gradients, provide visual explanations by attributing importance to input pixels. Unfortunately, they may suffer from instability with even slightly perturbed inputs. Layer-wise Relevance Propagation (LRP[3]) attempts to address some of these issues by backpropagating relevance scores through the network layers. While LRP offers theoretically sound attributions, it is highly sensitive to the choice of propagation rules and often requires architecture-specific tuning, which limits its practical applications. Prototype-based methods such as ProtoPNet[4] offer more intuitive visual explanations by linking predictions to learned prototypes, but do so at the cost of reduced flexibility and model capacity. Methods such as SHAP[5] and LIME[6] are commonly used for tabular and sequential data, relying on surrogate models and perturbations to estimate feature importance, while unfortunately also introducing potential approximation errors. Except for LRP, all the above techniques are applied externally, *post hoc*, to a pre-trained model. By contrast, DxANN embeds interpretability into the model training, *ante hoc*, and produces per-sample, per-feature explanations natively and efficiently, with a natural applicability to images and a potential for extension to other data modalities.

## 3. Theoretical background

DxANN is based on the Real-NVP (Real-valued Non-Volume Preserving) algorithm—a flow-based generative model that maps complex data distributions to a simpler (typically Gaussian) distribution, using a sequence of computation-efficient bijective transformations of the form $y = s \odot x + t$ where s and t are learnable scale and translation parameters. The bijective transformations ATB are shown in Figure 1. The Real-NVP is unsupervised and is trained using the following loss function:

$$Loss = -log(q(z)) - log \sum_{k=1}^{K} |s_k|$$

where z is the latent representation of input x, K is the number of transformations, and q is a predefined (typically Gaussian) distribution. The term $s_k$ is the scaling factor of transformation k.

---

[2] Selvaraj, Ramprasad R., et al. "Grad-CAM: visual explanations from deep networks via gradient-based localization." International journal of computer vision 128 (2020): 336-359.

[3] Bach, Sebastian, et al. "On pixel-wise explanations for non-linear classifier decisions by layer-wise relevance propagation." PloS one 10.7 (2015): e0130140.

[4] Chen, Chaofan, et al. "This looks like that: deep learning for interpretable image recognition." Advances in neural information processing systems 32 (2019).

[5] Lundberg, Scott M., and Su-In Lee. "A unified approach to interpreting model predictions." Advances in neural information processing systems 30 (2017).

[6] Ribeiro, Marco Tulio, Sameer Singh, and Carlos Guestrin. "" Why should i trust you?" Explaining the predictions of any classifier." Proceedings of the 22nd ACM SIGKDD international conference on knowledge discovery and data mining. 2016.

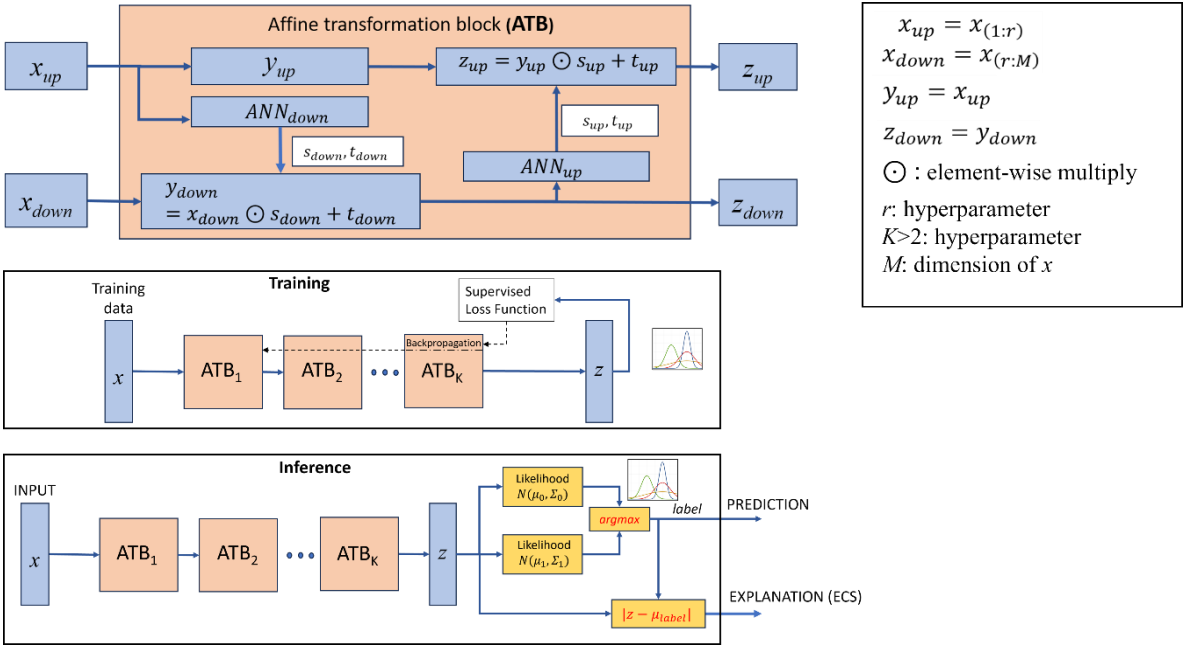

*Figure 1  The DxANN algorithm*

DxANN implements a binary classifier using the same flow as the Real-NVP, but instead of using one predefined normal distribution, two distributions are assigned, one for each of the two classification labels, with different means: $q_{z_0}(z) \sim N(\mu_0, I)$ and $q_{z_1}(z) \sim N(\mu_1, I)$. The resulting loss function is:

$$Loss = -y^{(n)} \cdot \left(log\left(q_{z_1}(z^{(n)})\right)\right) - (1 - y^{(n)}) \cdot \left(log\left(q_{z_0}(z^{(n)})\right)\right) - \sum_{k=1}^{K} log\left(\left|S_k^{(n)}\right|\right)$$

Where $y^{(n)}$ is the label of the $n^{th}$ sample and $z^{(n)}$ is the embedding of the $n^{th}$ sample in the latent z space. After training the model, the classification inference is determined by selecting the distribution that maximizes the embedded sample's likelihood:

$$prediction = argmax\left(q_{z_0}(z^n), q_{z_1}(z^n)\right)$$

To attain per-feature explainability of the inferences, we use Explainability Contribution Score (ECS) - a real-valued vector metric whose components reflect the relative importance of the corresponding components of the input data sample in generating the inference. ECS is particularly effective for image data but generalizes naturally to other structured inputs.

Let $z_{nm}$ denote the embedded representation of feature *m* of input sample *n*, and let $\mu_i$ represent the mean of the distribution assigned to class *i* in the latent space z. The ECS is defined as:

$$ECS_{nm} \sim |z_{nm} - \mu_i|$$

The ECS is therefore proportional to the distance between the embedded feature and the mean of its embedded class distribution. In essence, it captures how "atypical" a feature is within its class-specific embedding distribution. Intuitively, the ECS reflects the difficulty faced by the model in embedding a given feature in a way that aligns with the learned embedded representation for the class. For images, assuming that each pixel is treated as a separate feature, a feature gains importance only by virtue of its spatial context and relation to the neighboring features (pixels), and a high ECS value signals that the model had to make a more significant difficulty to reconcile the pixel with the expected class— implying a higher contribution to the classification decision.

# 4. Experimental setup and results

We tested DxANN on two labeled datasets sourced from Kaggle and Mendeley Data. The first dataset[7] contained 4029 OCT images of retinas of patients with and without Diabetic Macular Edema (DME). The second dataset[8] contained 3988 images of knees of patients with and without osteoarthritis (OA). Each of the two datasets was randomly split 80% / 20% to training/test sets. The binary classification task was diagnosis of DME and OA, respectively.

The ANNs in Figure 1 can be any of deep neural network type. For images, we experimented with CNN as the ANNs. The CNN-based DxANN model was trained on the above two labeled datasets. The models included two ATBs (see Figure 1), each containing eight convolution layers and one hidden, fully connected layer. The total number of parameters was ~13 million. The training set was divided into 64 batches, each containing ~50 samples, and the model trained for ~2000 epochs.

To compare the accuracy of the DxANN-based diagnostic models to state of the art foundation models, we pre-trained the ResNet-50[9] [10] and the VGG-16[11] [12] - based foundation models on the ILSVRC 2012 ImageNet Large Scale Visual Recognition Challenge dataset, and then fine-tuned them for each of the two medical diagnosis tasks by unfreezing the last four layers and training them along with a custom dense classification head. A dense layer with 256 neurons and ReLU activation was added, followed by a dropout layer with a rate of 0.5. The final dense layer used a softmax activation function. The training was performed using Adam optimizer with a learning rate of $0.0001$. The model was trained for 20 epochs with a batch size of 32.

In addition, to compare the accuracy of the CNN-based DxANN with that of a traditional opaque task-specific CNN, we also implemented a custom CNN classifier trained (like DxANN) from scratch on the same two labeled DME and OA image datasets.

**Accuracy.** The performance of the model was evaluated in terms of prediction accuracy and explainability. The results are summarized in the table below, which shows that accuracy of the CNN-based DxANN is comparable to that of state of the art foundation model-based classifiers, and matches or exceeds the accuracy of a plain task-optimized CNN:

| Algorithm | DME diagnosis accuracy | OA diagnosis accuracy | Explainability |
|---|---|---|---|
| **Custom designed CNN** | 93.7% | 94.4% | None |
| **Fine-tuned VGG-16** | 97.6% | 97.6% | None |
| **Fine-tuned ResNet-50** | 96.3% | 96.7% | None |
| **CNN-based DxANN** | **97.1%** | **97.2%** | ***Deep, per sample, per feature*** |

**Explainability.** Figure 2 shows the explainability outputs generated by the CNN-based DxANN model in two examples for each of the two medical diagnosis tasks, in the form of ECS-based heatmap overlays applied to the original images. For retinal OCT images, the model highlights regions associated with pathological changes such as retinal swelling or fluid accumulation. In the knee x-ray images, the model highlights structural regions of diagnostic relevance — including joint boundaries and potential degenerative regions - thus providing the user with cues that explain the diagnosis. The explanation heatmaps are colored according to the ECS assigned by the model to each pixel, with brighter yellow color assigned to the more influential pixels, and darker red to the less influential. The correspondence of the resulting visual explanations with standard clinical insights was confirmed in our informal presentations to ophthalmologists and orthopedists, respectively.

---

[7] Kaggle, Retinal OCT Images dataset

[8] Mendeley Data, Knee Osteoarthritis Severity Grading Dataset

[9] He, Kaiming, et al. "Deep residual learning for image recognition." Proceedings of the IEEE conference on computer vision and pattern recognition. 2016.

[10] https://www.tensorflow.org/api_docs/python/tf/keras/applications/ResNet50

[11] Simonyan, Karen, and Andrew Zisserman. "Very deep convolutional networks for large-scale image recognition." arXiv preprint arXiv:1409.1556 (2014).

[12] https://www.tensorflow.org/api_docs/python/tf/keras/applications/VGG16

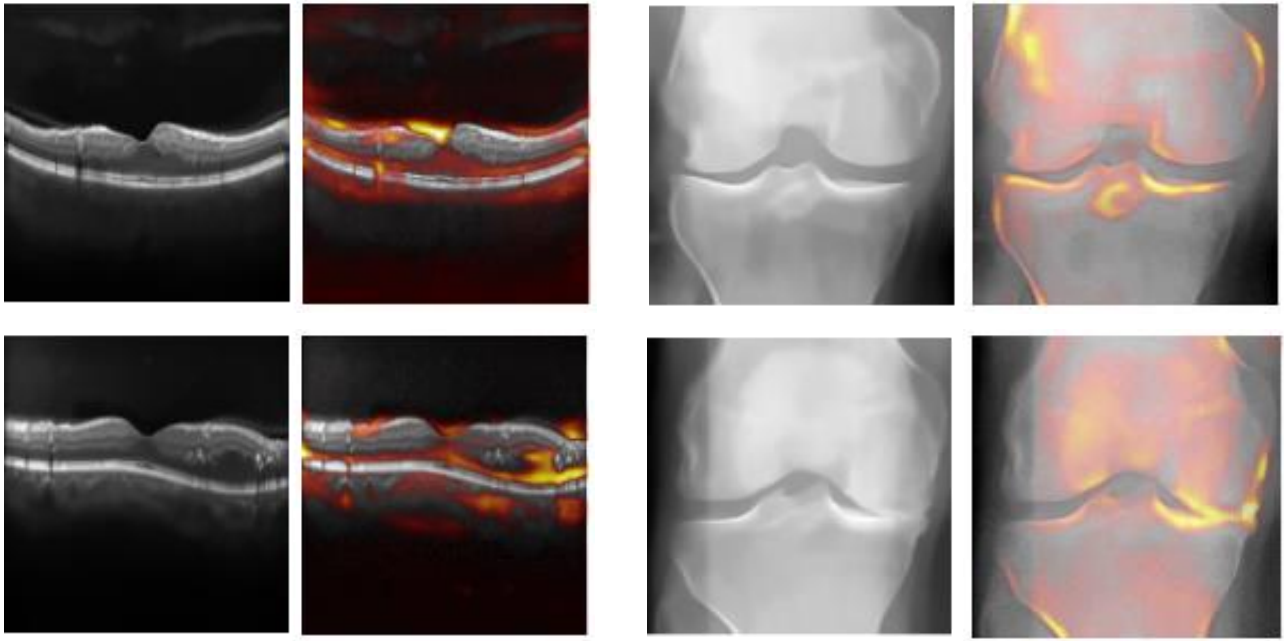

(left) OCT images of retinas correctly diagnosed with DME by DxANN; (right) DxANN's explanation of the diagnosis, with brighter colors pointing out the most influential areas,

(left) X-ray images of knees correctly diagnosed with OA by DxANN; (right) DxANN's explanation of the diagnosis, with brighter colors pointing out the most influential areas.

*Figure 2*

## 5. Discussion

The growing demand for explainable artificial intelligence (XAI) in medical imaging reflects the need for models that not only perform well but also offer transparent, clinically meaningful insights. The proposed DxANN method and architecture meet both of these objectives by combining a high classification performance with an inherent, per-sample, per-feature explainability. Unlike the *post hoc* tools such as SHAP or Grad-CAM, DxANN's Explainability Contribution Score (ECS) is computed directly in the model's latent space and does not rely on approximations or perturbations. By integrating the *ante hoc* explainability into the architecture of the model, we obtain explanations that reflect the model's internal reasoning.

In our experiments, the CNN-based DxANN model consistently highlighted the regions for retinal OCT and musculoskeletal scans that were relevant to its diagnostic inferences, while performing the diagnosis with a very high accuracy. This demonstrates its ability to prioritize regions that align with clinical intuition. Importantly, when each feature is a pixel, the relevance of individual pixels to the inference depends only on the spatial context and relation between the pixels. DxANN captures this dependence by measuring the distance between each embedded feature and its class-conditional mean in the latent space, treating larger deviations as indicators of stronger influence on the inference.

## 6. Conclusions

We presented DxANN, a novel self-explaining deep learning classifier, and reported test results for two image-based medical diagnosis tasks. Central to our approach is the Explainability Contribution Score (ECS), a metric derived from latent feature representations that reflects the influence of individual pixels on the model´s inference. Unlike traditional explainability methods, DxANN provides intrinsic explainability and does not require external tools or approximation methods.

Our results show that DxANN not only matches the learning performance of the widely used ResNet-50 and VGG-16 based foundation models fine-tuned for specific tasks, but also provides rich and clinically aligned explanations. Future work will explore extending DxANN and its explainability metrics to other data modalities, multimodal inputs and other model architectures such as MLP, transformers and RNNs.